\documentclass{article} 
\usepackage{nips10submit_e,times}
\usepackage[dvips]{graphicx}
\usepackage{amssymb,amsmath}
\usepackage{nccmath}

\title{The thermodynamic temperature of a rhythmic spiking network}

\author{
Paul Merolla\thanks{pmerolla@gmail.com} \\
Stanford University\\
\And
Tristan Ursell \\
Stanford University \\
\texttt{} \\
\AND
John Arthur \\
Stanford University \\
}

\newcommand{\insfig}[5]{
\begin{figure}[#5]
    \begin{center}
        \scalebox{#2}{\includegraphics{#1.eps}}
    \end{center}
    \caption{#3}{\sffamily \small #4}
    \label{fig:#1}
\end{figure}
}

\newcommand{\inssfig}[7]{
\begin{figure}[#7]
  \begin{center}
        \begin{tabular}{cc}
            \scalebox{#2}{\includegraphics{#1.eps}} &
            \scalebox{#4}{\includegraphics{#3.eps}} \\
            (a) & (b)
        \end{tabular}
    \end{center}
    \caption{#5}{\sffamily \small #6}
    \label{fig:#1}
\end{figure}
}

\newcommand{\inslfig}[9]{
\begin{figure}[#9]
    \begin{center}
        \begin{tabular}{ccc}
            \scalebox{#2}{\includegraphics{#1.eps}} &
            \scalebox{#4}{\includegraphics{#3.eps}} &
            \scalebox{#6}{\includegraphics{#5.eps}} \\
            (a) & (b) & (c)
        \end{tabular}
   \end{center}
    \caption{#7}{\sffamily \small #8}
    \label{fig:#1}
\end{figure}
}

\newcommand{\insxxfig}[9]{
\begin{figure}[#9]
    \begin{center}
        \begin{tabular}{cc}
            \scalebox{#2}{\includegraphics{#1.eps}} &
            \scalebox{#4}{\includegraphics{#3.eps}} \\
	   (a) & (b)\\
	   \multicolumn{2}{c}{\scalebox{#6}{\includegraphics{#5.eps}}} \\
            \multicolumn{2}{c}{(c)}
        \end{tabular}
   \end{center}
    \caption{#7}{\sffamily \small #8}
    \label{fig:#1}
\end{figure}
}

\nipsfinalcopy 

\begin{document}

\maketitle

\begin{abstract}
Artificial neural networks built from two-state neurons are powerful computational substrates, whose computational ability is well understood by analogy with statistical mechanics.  In this work, we introduce similar analogies in the context of spiking neurons in a fixed time window, where excitatory and inhibitory inputs drawn from a Poisson distribution play the role of temperature.  For single neurons with a ``bandgap" between their inputs and the spike threshold, this temperature allows for stochastic spiking.  By imposing a global inhibitory rhythm over the fixed time windows, we connect neurons into a network that exhibits synchronous, clock-like updating akin to neural networks.  We implement a single-layer Boltzmann machine without learning to demonstrate our model.  
\end{abstract}

\section{Introduction}
How networks of spiking neurons compute in the presence of noise is a topic of active research~\cite{Rao05hierarchicalbayesian,Deneve,Ma2006Bayesian}.  One intriguing possibility is that the noise plays an important role, similar to how stochasticity in artificial neural networks is used to solve a wide-range of problems, from recognition to optimization~\cite{haykin-1994}.  However, despite initial progress~\cite{Hinton98spikingboltzmann}, there is still no systematic way to implement stochastic algorithms commonly found in neural networks using spiking neurons.

The fundamental problem is that the mathematical framework of artificial neural networks, which has deep ties to statistical mechanics, lacks a meaningful definition of temperature in the presence of dynamically spiking neurons.  For example, the dynamics of the well-known Hopfield network are governed by an energy function that converges to a set of stable attractors~\cite{JJHopfield82}.  However, when two-state neurons in the Hopfield network are replaced by stochastically spiking neurons, the network exhibits non-equilibrium dynamics.  In other words, spiking networks do not reach a thermal equilibrium, and hence the link between network dynamics and statistical mechanics is broken~\cite{Bialek}. 

In this paper, we describe a methodology for mapping artificial neural networks onto a network of stochastic spiking neurons.  In particular, we present an architecture where Poisson input noise sets the thermodynamic ``temperature" of the network.  Our approach is twofold: First, we define the notions of energy and temperature for an integrate-and-fire neuron.  The insight here is to set a fixed time window for the neuron to spike, analogous to a discrete time step in an artificial neural network.  Second, we construct a network and impose a global inhibitory rhythm that restricts each neuron to spike within a synchronized time window, thereby ensuring measures of energy and temperature are consistent across the network.  Based on this algorithm, it is possible to implement a range of neural network algorithms (e.g., the Boltzmann machine~\cite{Hinton2006}) built from neurons that spike.


\section{Thermodynamic equivalents for spiking neurons}

To arrive at thermodynamic measures in the context of spikes, our strategy is to find a regime where the behavior of a two state neuron from artificial neural networks can be approximated by a spiking neuron.  


\subsection{Bandgap of a I\&F neuron}
\label{sec:Energy}

In artificial neural networks, the state of an abstract two-state neuron, $s_i$, with an input $x$ and threshold $x_t$ is set by
\\ \[
s_i = \left\{\begin{array}{ll}
        1 & \mbox {for } x>x_t, \\
        0 & \mbox {for } x \leq x_t\end{array}\right. 
\] \\
for each discrete time step.  We refer to the above neuron model as a ``two-state" neuron.  For subthreshold input, the ``bandgap" ($e_g$) is defined as the additional input required to switch states, given by $e_g=x_t - x$. 

To introduce the notion of a bandgap for a neuron that spikes, consider the dynamics of a leaky integrate-and-fire model (I\&F) 
\\ \[
\tau_m \dot{u}(t) = u_o-u(t) + R(I_{in}(t) - I_\alpha(t)),
\] \\
where $\tau_m$ is the membrane time constant, $u(t)$ is the membrane potential, $I_{in}(t)$ is the input current, $I_\alpha(t)$ is a spike adaptation current, and $R$ is the input resistance.  For each instance that $u$ exceeds a threshold $u_t$, the neuron generates a spike denoted by times $t^{(f)}$, after which the membrane is reset to $u_o$.  Each spike also causes the spike adaptation current to increase as a $\delta$-function impulse with magnitude $\Delta_{\alpha}$, and decays according to 
\\ \[
\tau_\alpha \dot{I_\alpha}(t)=-I_\alpha(t)+\Delta_{\alpha} \delta(t-t^{(f)}), 
\] \\
where $\tau_\alpha$ is the adaptation time constant.  

This spiking model mimics the behavior of a two-state neuron under the following constraints: 
\begin{enumerate}
\item 
The opportunity for spiking is restricted to a fixed time window $T_W$ where $u$ has the initial condition $u_o$. 
\item
The neuron's membrane time constant is short compared to the fixed time window ($\tau_m \ll T_W$).
\item 
The output of the neuron is cast into a binary variable (``on" or ``off"), defined as the event that the neuron spikes at least once in the window.  
\item 
The input is the fixed value $I_o$ within $T_W$.
\end{enumerate}

\inslfig{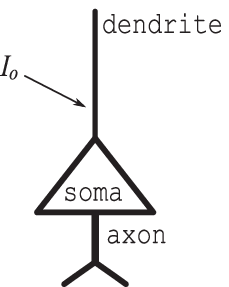}{1}{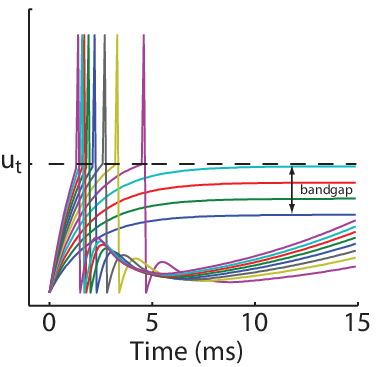}{1}{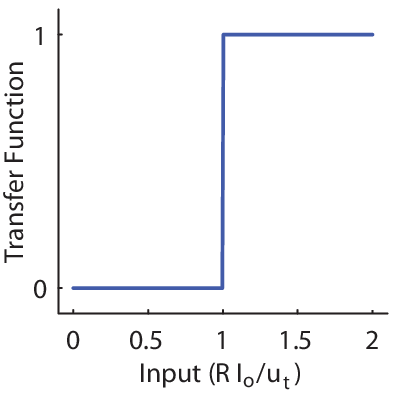}{1}{I\&F as a two-state neuron}{(a) A point I\&F neuron that receives direct input current $I_o$. (b) Membrane potential response for different levels of $I_o$ ranging from 0.6 to 2.  In this simulation, $\tau_m=2$ ms, $R=1$, $T_W=15$ ms, and $u_t=$1.  The membrane follows the exponential trajectory: $u(t) = R I_o(1-e^{(-\frac{t}{\tau_m})})$ that asymptotes to $R I_o$ when subthreshold; if the membrane crosses threshold (dotted line), the potential jumps to $2u_t$ (spikes), at which point it is reset back to 0; note that in these simulations, our model also includes spike adaptation to prevent the neuron from spiking again within the window.  (c) The neuron's transfer function is a step with a sharp transition at 1.}{t}

Applying these constraints, the I\&F neuron elicits a spike (``on" state) or not (``off" state) based on the level of input current $I_o$.  In Figure~\ref{fig:Neuron0}b, the different traces represent the response of the membrane potential to different inputs, applied in independent trials.  The transfer function, which links the neuron's state to its input strength, is identical to that of a two-state neuron, where a critical value of input is required to spike (Figure~\ref{fig:Neuron0}c). 

By analogy to the two-state neuron, we denote the bandgap of the I\&F neuron as the additional input required to spike in $T_W$, which corresponds to $e_g\simeq u_t - R I_o$.  Note that because the exponential trajectory saturates quickly to $R I_o$ (after $3\tau_m$), the instantaneous distance to threshold is approximately fixed (see Figure~\ref{fig:Neuron0}b).  Next, we demonstrate that adding noisy inputs to the I\&F neuron leads to stochastic spiking---analogous to how adding a non-zero temperature to a two-state neuron leads to stochastic flipping of states.


\subsection{Non-zero temperature by including Poisson spikes}

To add a non-zero temperature to a two-state neuron, a standard method is to flip its state with a probability according to the logistic function
\\ \[
P = \frac{1}{1 + e^{-\frac{e_g}{T}}}
\] \\
where $e_g$ is the bandgap and $T$ is the temperature~\cite{Ackley85alearning}.  We are able to capture this same basic relationship in an I\&F neuron by driving it---in addition to the constant drive that it already receives---with spikes drawn from Poisson statistics.  The intuition is that a neuron which would otherwise not spike in the fixed time window now has the chance to cross threshold due to the stochastic inputs.  The odds of stochastically crossing the threshold increase as the bandgap decreases. 

Consider the stochastic arrival of excitatory and inhibitory inputs that impart fixed charges $w_e$ and $w_i$ respectively, as $\delta$-function impulses with Poisson rates $\lambda_E$ and $\lambda_I$.  The input current is modified to
\\ \[
I_{in}(t) = I_o + w_e\sum_{j}\delta(t-t_j^{(E)}) - w_i\sum_{k}\delta(t-t_k^{(I)}),
\] \\
with the arrival times $t_j^{(E)}$ and $t_k^{(I)}$ determined stochastically from their respective Poisson distributions.

\inslfig{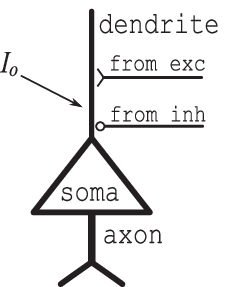}{1}{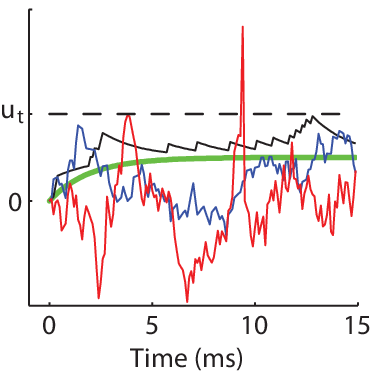}{1}{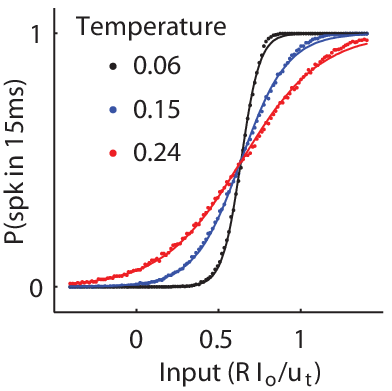}{1}{I\&F neuron with noise}{(a) I\&F neuron that receives a constant input current $I_o$, along with noisy spike trains from excitatory and inhibitory populations with mean rates $\lambda_E$ and $\lambda_I$, respectively. (b) Sample paths of I\&F membrane potential when $w_e=w_i=0.1$, and $I_o=0.5$ for: no noise (thick green), low noise (black), mid noise (blue), and high noise (red).  (c) Measured neuron transfer function for the same three levels of noise.  The effective temperature is extracted by fitting the measured sigmoid (points) to a logistic function (lines).}{t}

We simulated the dynamics of an I\&F neuron with noise for three Poisson rates (in Hz): $\lambda_E=$920 and $\lambda_I=$0 (low-noise regime); $\lambda_E=$5,000 and $\lambda_I=$6,150 (mid-noise regime); and $\lambda_E=$14,000 and $\lambda_I=$17,500 (high-noise regime).\footnote{These rates represent the combined effect of $N$ synaptic inputs times the average rate of synaptic events.}  Sample paths for $u(t)$ are shown for the three noise settings in Figure~\ref{fig:Neuron1}b when $I_o=0.5$ for a fixed threshold (black, blue, and red, respectively); for reference, we also show the response with no noise (thick green).  As expected, the membrane potential variance, and hence the odds of crossing threshold, increases with the level of noise.  The neuron's transfer function is well fit by the logistic function where the slope of the sigmoid becomes less steep at higher noise levels (Figure~\ref{fig:Neuron1}c), and the corresponding fits for the effective temperature are 0.06, 0.15, and 0.24 for low, mid, and high noise respectively.

\subsection{Formula for temperature}

We derive an analytic approximation for temperature in this model as follows: The first assumption is that the neuron can only spike after $\gamma \tau_m$, the region where $u_t - R I_o$ is approximately constant.  We choose $\gamma =$3 by inspection.\footnote{Note that $\gamma$ is the only free parameter that we have in our analysis.}  Next we assume that $u$ follows an Ornstein-Uhlenbeck process with mean membrane potential $\mu = R I_0 + \tau_m (\lambda_e w_e - \lambda_i w_i)$ and noise amplitude $\sigma^2 = \tau_m (\lambda_e w_e^2 +\lambda_i w_i^2 )$; in the regime where spike magnitudes $w_e$ and $w_i$ are small and their respective arrival rates are large, this is a good approximation~\cite{Gerstner}.  Finally, we approximate the distribution of first-passage times to cross the threshold, which in general is a difficult distribution to find analytically, with an exponential probability density function that depends only on the mean first passage-time (MFPT).  This approximation is supported by our empirical analysis (data not shown), and also has theoretical support~\cite{Ricciardi}.  


Under these assumptions, the probability to spike as a function of $\mu$ is given by the cumulative distribution of first-passage times
\\ \[
\label{eq:pspk}
P_{spk}(\mu) = 1-e^{-T^{\prime}_W/T_\mu},
\] \\
where $T^{\prime}_W = T_W - \gamma \tau_m$.  $T_\mu$ is the MFPT with the initial condition $u_o=\mu$, given by
\\ \[
\label{eq:tmu}
T_\mu = \tau_m \int_0^{\frac{u_{t}-\mu}{\sigma}} \! f(x) \, dx,
\] \\
where $f(x) = \sqrt{\pi} e^{x^2}\left(1 + \mbox{erf }(x) \right)$ (see~\cite{Brunel99dynamicsof}).  We arrive at an approximation for temperature by computing the slope, $m$, at $P_{spk} = 1/2$, given by
\\ \[
m = \frac{1}{2} \left(\log{2}\right)^2 \frac{\tau_m}{T^{\prime}_W\sigma} f\left(\frac{u_{t}-\mu}{\sigma}\right),
\] \\
and equating this slope with the midpoint slope of the logistic function, $1/4T$.  Solving for temperature yields
\\ \[
T = \frac{T^{\prime}_W \sigma}{2 \left(\log{2}\right)^2 {\tau_m} f\left(\frac{u_{t}-\mu}{\sigma}\right)}.
\] \\
Using the values for low, mid, and high noise regimes, our temperature approximations (0.04, 0.15, and 0.25 respectively) closely match the simulation (see Figure~\ref{fig:Neuron1}c).

%


\section{Implementing a connected network}



When neurons are allowed to spike beyond the confines of a fixed time window, our measure for temperature breaks down.  Furthermore, it is not clear how to connect multiple neurons together into a network.  Here, we address these issues by imposing order on neuron dynamics using a global inhibitory rhythm. 

\subsection{Clocking neurons with an inhibitory rhythm}
\label{sec:Clock}


The idea of adding a separate neural control structure that essentially ``clocks" network activity, was explored by Menschik et. al. in the context of a hippocampal model for recalling memories~\cite{Menschik}.  The main insight is that a strong inhibitory rhythm forces spiking neurons to operate as if there are discrete time steps.  For example, a 33Hz square-wave rhythm applied to an I\&F neuron restricts the opportunity to spike to a 15 ms window, interspersed by periods of high inhibition when no spikes can occur.  In essence, one can think of the periods when spiking is allowed as discrete updates when the neuron is evaluating its inputs, and the inhibitory periods as reset phases.  

\insfig{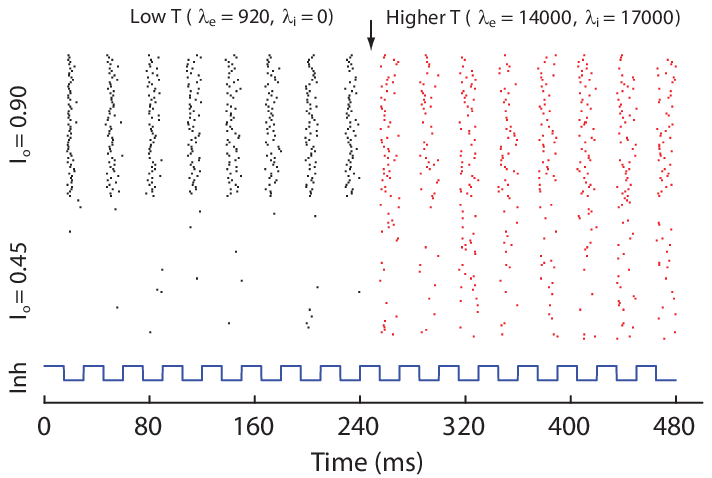}{1}{Changing network temperature dynamically}{A population of 128 neurons arranged from top to bottom receives a 33Hz square-wave inhibitory rhythm (Inh, blue trace), which enforces the constraint that spikes only occur in the low inhibition phase, and noisy excitatory and inhibitory inputs (not shown).  Each spike event is indicated by a vertical dash at the neuron's location ($y$ dimension) and its time ($x$ dimension), to create a raster plot.  The top 64 neurons are driven with an input of 0.9, whereas the bottom 64 are driven with 0.45.  At 255ms (arrow), the level of background noise is changed from low to high, representing a change in temperature from 0.06 to 0.24.}{t}

Consider the dynamics of an unconnected network of 128 I\&F neurons, where neurons in the top half are driven by high input current, and neurons in the bottom half are driven by low input current (Figure~\ref{fig:raster}).  In addition to this constant input, each neuron receives a strong inhibitory rhythm to clock network dynamics, as well as excitatory and inhibitory spikes with Poisson statistics to set the effective temperature. 

In Figure~\ref{fig:raster} we demonstrate the effect of changing the temperature of this network when it is running continuously.  For the first 255 ms, the temperature of the network is set to 0.06.  Neurons in this first epoch spike mostly in a deterministic manner (black dashes).  Specifically, at high input (top half), most neurons spike in each of the 8 phases of the rhythm, while at low input (bottom half), most neurons do not spike.  After 255 ms, we abruptly raise the temperature to 0.24 by increasing the level of Poisson noise.  In this second epoch we observe that neurons respond more stochastically (red dashes).  On average, the odds of spiking to the high input are slightly lower than before, yet the neurons are more sensitive to lower inputs.  In a sense, we have increased the dynamic range of the neuron simply by increasing its temperature; the price to pay, however, is that neurons now spike less deterministically (i.e., they may spike in the absence of input, or conversely, they may not spike in its presence).


\subsection{Linking network states across time}
\label{sec:Markov}

To implement a connected network, we need to determine how outputs (spikes) are passed along as inputs (currents) from local connections within the network.  Our strategy is to specify a synaptic waveform, $\alpha(t)$, that allows the dynamics of the spiking network to exactly match the synchronous update of an artificial neural network.  Unlike the Poisson inputs described earlier, these synaptic inputs are generated in a deterministic manner (similar to how the input to a two-state neuron is generated deterministically, before applying the logistic function and comparing to a random number drawn from a uniform distribution).



In a fully connected artificial neural network, the input to the $j$th neuron in time step $k$ is written as $x_j(k) = \medop\sum_p w_{jp} s_p(k-1)$, with $w_{jp}$ as the connection strength between neuron $j$ and $p$, and $s_p(k-1)$ is the binary value of neuron $p$ from the previous step.  This update rule can be mapped to a network of I\&F neurons as follows: First, the input to each neuron, $I_o$, is mapped to $x$ for each discrete time step.  Next, the discrete time step $k$ maps to the time period $T_W+2 T_W(k-1)<t<2T_W k$, corresponding to the $k$th low phase of the inhibitory rhythm.  Then, $s_p(k)$ is taken to be the binary event that neuron $p$ spikes in $k$th phase.  Finally, the postsynaptic current at neuron $j$ induced by a presynaptic spike from neuron $p$ is a delayed pulse given by
\\ \[
\alpha(t)_{jp} = w_{jp}\Theta(t-t_p^{(f)}+T_W)(1-\Theta(t-t_p^{(f)}+3 T_W)),
\] \\
where $\Theta(t)$ is the Heaviside step function.  The first step (left term) models a fixed axonal delay of $T_W$, and ensures that spikes in the current window do not influence targets in this same window.  The second step (right term) sets the length of the pulse at $2 T_W$, and ensures that its effect lasts the full duration of the following phase.  Note that the form of $\alpha(t)$ was chosen to exactly match the neural networks case---variations of $\alpha(t)$ are described in the discussion.  
 
%
 

\inssfig{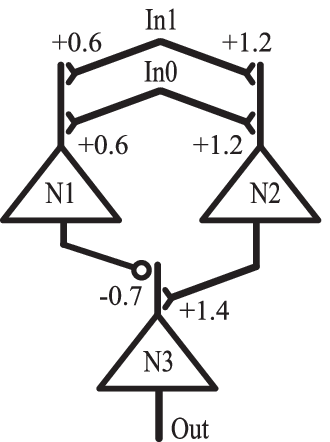}{1}{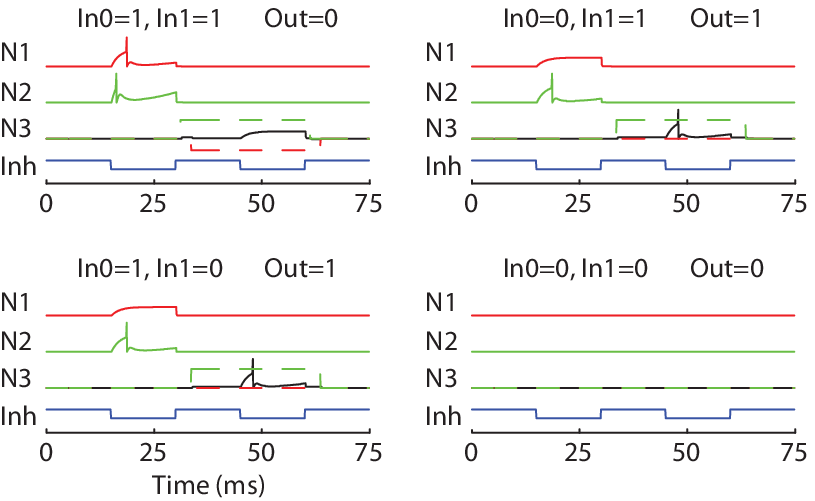}{1}{{\sc XOR} with spikes}{(a) A two-layer feed-forward network at zero temperature that implements an {\sc XOR} function.  Neurons N1--N2 are driven by external binary-valued inputs In0 and In1 for 30 ms with indicated weights.  N1 in turn inhibits N3, while N2 excites N3.  All neurons receive global rhythmic inhibition (connections not shown).  (b) Neuron responses for all input combinations, and the resulting postsynaptic currents induced at N3 (dashed lines).  N1--N2 evaluate their inputs in the low phase of the inhibitory rhythm (15 to 30 ms), and their spikes generate a delayed step response at N3 that lasts the full duration of the second phase (45 to 60 ms).  N3 crosses threshold when N2 was active and N1 was not active in the previous phase, thus performing an {\sc XOR}.}{t}




With these choices, one can use spiking neurons to mimic the dynamics of an interconnected network of two-state neurons.  For a simple demonstration, we map the well known {\sc XOR} problem to a feed-forward network of three I\&F neurons (Figure~\ref{fig:XOR}).  In the following section, we show how to implement a stochastic recurrent network, known as the Boltzmann machine.

\section{A one-layer Boltzmann machine}
\label{sec:BMach}

The Boltzmann machine, introduced by Ackley et. al. in 1985, is a stochastic neural network that has been used to perform recognition, as well as learn data representations without supervision~\cite{Hinton2006}.  For simplicity, we implement a fully-connected one-layer Boltzmann machine without learning using spiking neurons.  However, it is worth noting that the Boltzmann machine has a local learning rule that is ripe for implementation using spike-based mechanisms (e.g., spike-timing dependent-plasticity).  Our primary goal is to demonstrate that the network's temperature, which we set by the level of Poisson noise, leads to stochastic flipping between attractor states.  

The Boltzmann machine algorithm that we consider has N-connected two-state neurons $s_1, s_2, \ldots, s_N$ such that the state of neuron $i$ flips on with probability
\\ \[
p_i = \frac{1}{1 + \mathsf{exp}(-\medop\sum_{j} w_{ij}s_j /T)},
\] \\
where $T$ is the thermodynamic temperature.  As is standard for the Boltzmann machine, we remove self-connections ($w_{ii}=0$ for all $i$), and require symmetry ($w_{ij}=w_{ji}$ for all $i,j$).  By applying the ideas developed in this paper, we are able to map the above algorithm onto a recurrent network of spiking neurons (Figure~\ref{fig:Network5}a).  



To test the mapping, we simulate a network of 128 recurrently-connected I\&F neurons using a 33Hz inhibitory rhythm and a temperature of 0.24.  The weight matrix $w_{ij}$ is constructed according to the method used in~\cite{JJHopfield82}.  Specifically, we choose four random binary attractors, $V_s$, each of length N and 75\% sparsity, and compute $w_{ij}=\medop\sum_{s} (2V_i^s-1)(2V_j^s-1)$.  Note that attractors are chosen to overlap with at least one other attractor by at least 30\%.


\insxxfig{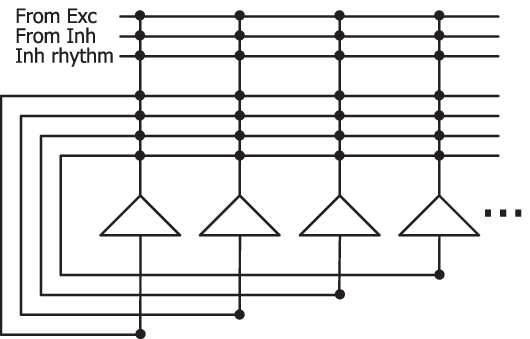}{1}{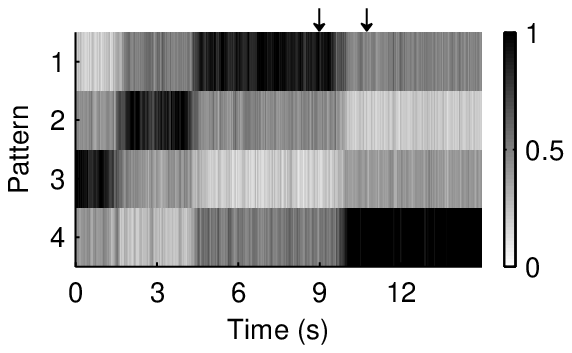}{1}{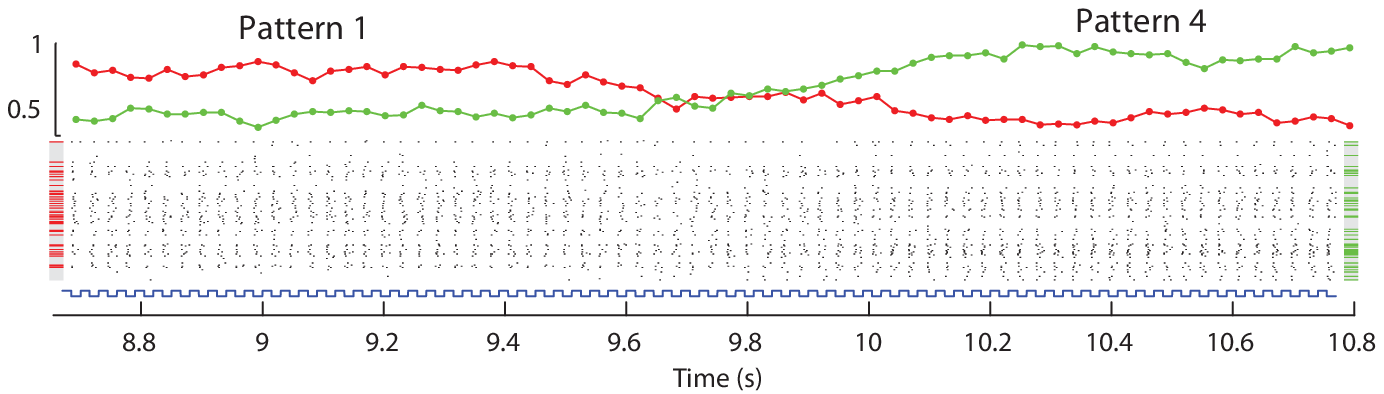}{1}{Boltzmann Machine with I\&F neurons}{(a) A recurrent network of 128 I\&F neurons that receives: noisy inputs from excitatory and inhibitory populations, global rhythmic inhibition, and recurrent inputs.  (b) The correlation of the network's binary state fluctuates stochastically between four attractors.  (c) The detailed network trajectory from attractor 1 (red, left) to attractor 4 (green, right), plotted as a spike a raster over multiple inhibitory cycles.}{t}

We explored the attractor dynamics by correlating binary-valued states $S_V(k) = \{s_i \ldots s_N \}$ in cycle $k$ with the four stored patterns $V^s$ by taking the inner product.  Indeed, the correlation closely matched one of the four patterns for the majority of the simulation (Figure~\ref{fig:Network5}b).  On occasion, the network abruptly transitions between states due to the network's non-zero temperature---a hallmark trait of a Boltzmann machine with discrete attractors.  Figure 5c shows the detailed dynamic transition between states 1 and 4 within a specified time window.


\section{Discussion}

We presented a recipe for implementing common neural network algorithms using spiking neurons.  We build on previous work~\cite{Menschik} by making explicit parallels to statistical mechanics.  To our knowledge, this is the first time a measure of thermodynamic temperature has been linked to neurons that spike probabilistically.    

We view this work as a first step to bringing more biological relevance and physicality to abstract two-state neurons.  However, as currently presented, the details of our model are biologically unnatural.  

The membrane time constant used in our simulations is relatively short (2 ms) when compared to membrane time constants typically observed in the cortex (10 to 20 ms).  However, if biological neurons are sufficiently leaky, their effective time constant can be quite short ($<$5 ms) even though the membrane time constant is longer~\cite{Gerstner}.  

More fundamentally, the short membrane time constant was chosen to ensure the membrane potential saturates quickly, leading to an intuitive definition for the bandgap.  For operation outside of the $\tau_m \ll T_W$ regime, the membrane potential does not saturate and confounds the measurement of network temperature.  We are able to define an effective bandgap for the window, however, further simulations are required to determine under what conditions a logistic function serves as a good approximation of the neuron transfer function.  However, in the case $\tau_m \gg T_W$ (when the time-dependent decay within a time step can be ignored), the transfer function is well known and closely resembles a logistic~\cite{Tuckwell}.  

To emulate discrete time steps in our network, we introduced a square wave inhibitory rhythm that is strong enough to reset each neuron's state at a regular interval.  Such a rhythm is unrealistic for a biological network in its present form.  For example, it is thought that brain rhythms are sinusoidal, and even though spikes can be entrained to the rhythm, they may occur at any phase~\cite{Singer}.  Although we did not explore these more biologically plausible rhythms, previous work from Menschik et al. demonstrates that a moderate-strength sinusoidal rhythm still leads to attractor-like dynamics in a recurrent network~\cite{Menschik}.

The synaptic waveform used in Section~\ref{sec:Markov} mimics a synchronous update in artificial neural networks.  This waveform was used only for clarity and is not the only choice.  For example, adopting a more biologically-relevant synapse with a rise and decay that lasts multiple inhibitory cycles, the network state could evolve in a more continuous and asynchronous manner~\cite{Yamanaka19971103}.  In this case, the current state of the network would depend on a handful of previous states each discounted by time, and we conjecture that this could provide a ``stiffness" against random network fluctuations.  



More pragmatically, this work could have implications for high performance computing.  Based on the present algorithm, one can implement many off-the-shelf neural network algorithms using neuromorphic architectures, which offer a scalable way to build large interconnected networks of spiking neurons~\cite{Merolla_07}.  Learning the synaptic connections in these large networks then becomes the bottleneck.  We are currently working to map local learning rules, such as contrastive divergence~\cite{HintonCD} to spiking networks using the same approach presented in this paper.





\bibliographystyle{plain}	
\bibliography{myrefs}		

\end{document}